\documentclass[11pt,a4paper]{article}
\usepackage[hyperref]{emnlp2020}
\usepackage{times}
\usepackage{latexsym}

\usepackage{graphicx}
\usepackage{algorithm,algpseudocode}
\usepackage{enumitem}
\usepackage{amsmath}
\usepackage[subtle,lists=normal]{savetrees}

\newcommand{\varParamA}{\varfont{Predict}}
\newcommand{\varParamB}{\varfont{Top10}}
\newcommand{\varParamC}{\varfont{List}}
\newcommand{\varParamD}{\varfont{Pop}}
\newcommand{\varParamE}{\varfont{str}}
\newcommand{\varParamF}{\varfont{Cat}}
\newcommand{\varParamG}{\varfont{ArgMax}}

\newcommand{\commentsymbol}{//}
\algrenewcommand\algorithmiccomment[1]{\hfill \commentsymbol{} #1}

\newcommand{\varfont}{\texttt}

\usepackage{microtype}

\aclfinalcopy 

\title{Are Some Words Worth More than Others?}

\author{Shiran Dudy\qquad Steven Bedrick\\
		Center for Spoken Language Understanding \\
	    Oregon Health \& Science University\\
	    Portland, Oregon, USA\\
	    {\tt \{dudy,bedricks\}@ohsu.edu}}

\date{}

\begin{document}

\maketitle
\begin{abstract}
Current evaluation metrics for language modeling and generation rely heavily on the accuracy of predicted (or generated) words as compared to a reference ground truth. 
While important, token-level accuracy only captures one aspect of a language model's behavior, and ignores linguistic properties of words that may allow some mis-predicted tokens to be useful in practice. 
Furthermore, statistics directly tied to prediction accuracy (including perplexity) may be confounded by the Zipfian nature of written language, as the majority of the prediction attempts will occur with frequently-occurring types.
A model's performance may vary greatly between high- and low-frequency words, which in practice could lead to failure modes such as repetitive and dull generated text being produced by a downstream consumer of a language model. 
To address this, we propose two new intrinsic evaluation measures within the framework of a simple word prediction task that are designed to give a more holistic picture of a language model's performance.
We evaluate several commonly-used large English language models using our proposed metrics, and demonstrate that our approach reveals functional differences in performance between the models that are obscured by more traditional metrics.

\end{abstract}

\section{Introduction}
\label{sec:intro}

Language models are foundational components in many NLP systems, and as such it is crucial to be able to empirically evaluate their behavior.
Traditionally, language models are evaluated using performance metrics that relate to the model's ability to accurately predict words given some context (e.g., perplexity).
Following the paradigm described by~\citet{galliers1993}, this can be thought of as an \emph{intrinsic} evaluation criterion (and perplexity an intrinsic metric), as it relates to the \emph{objective} of the language model itself.

In recent years, it has become common to also evaluate language models \emph{extrinsically}, in terms of the model's \emph{function}.
This is done by measuring a model's performance when used as a component in a downstream task.
\footnote{In part, this trend has been driven by the increasing use of downstream tasks as ancillary training objective functions; this somewhat confuses the traditional notion of intrinsic and extrinsic evaluation as a binary construct.} 
For example,~\citet{devlin-etal-2019-bert} evaluated BERT by using it as the language model component in benchmark tasks such as question answering and ``commonsense inference.''\footnote{SQuAD versions 1.1~\citep{rajpurkar-etal-2016-squad} and 2.0~\citep{rajpurkar-etal-2018-know}, and SWAG~\citep{zellers-etal-2018-swag}, respectively, in the case of the original BERT paper.}
This shift towards extrinsic and task-oriented evaluation is welcome, and has the potential to make language model evaluation more ecologically valid.
\footnote{
``Ecological validity'' is a dimension of experimental validity that is concerned with the question of whether an observed effect reflects ``what happens in everyday life''\citep{brewer_crano_2014}, i.e. beyond the artificial setting of the experiment itself. 
In an NLP context, a researcher working on question answering who was concerned with ecological validity would ensure that the questions on which they trained and evaluated their system were similar (in form and content) to those on which the system was designed to be used. }
As useful as task-oriented evaluation metrics are, however, we believe that this approach brings with it certain practical limitations, and that there remains a strong need for robust and meaningful intrinsic evaluation metrics that can be used to characterize and compare the performance of language models.

In this work, we outline and propose a variation on the standard next-word-prediction language modeling task that is designed for use in evaluating and comparing language models and is robust to implementation differences (tokenization method, etc.) that complicate the comparison of modern models in terms of token-level predictions. 
Our proposed metrics are richer and more meaningful measures than traditional intrinsic metrics such as perplexity, which is insensitive to \textit{which} tokens are matched, and as such may be confounded by distributional properties of their evaluation corpora. 
Our approach accounts not only for the \textit{accuracy} of a model's word predictions, but also the \textit{diversity of types} that it predicts, across different lexical frequency bins.
We further propose a formulation for the next-word-prediction task that explicitly allows for language- and task-level details to be captured in the resulting metrics, thereby blurring the line between intrinsic and extrinsic language model evaluation.
Our methods provide greater ecological validity than traditional intrinsic evaluation methods, while still remaining simple to interpret and easy to calculate. 

\subsection{Formalities: Language Models and Word Prediction}
\label{subsec:lm_def}

For our present purposes, we will consider a language model to be a model that, given a sequence $W$ of $n$ tokens $w_{1:n}$ from a fixed vocabulary of types $V$, estimates the joint probability of $P(W)$. 
The goal of a language models is of learning to approximate the distribution of tokens and types in some corpus.

Importantly, different models may use different units of prediction, at the level of individual character, at the word level, or (as with many modern neural models) at the level of a sub-word/sub-sentence unit (via e.g. byte-pair encoding~\citep{sennrich2016neural}, wordpieces~\citep{wu2016googles}, etc.).


Given such a model, we can typically also estimate the \emph{conditional} probability distribution $P(w_t | w_1 \ldots w_{t-1})$, over possible words occurring after a given history $h$ consisting of $t-1$ tokens. 
We refer to this as the next-word-prediction problem\footnote{Also known as the ``Shannon Game''~\citep{Shannon:1951}.} of predicting $\hat{w_t} = \operatorname*{argmax}_w P(w | h)$. Using the terminology of conditional text generation, this is akin to generating a single token via greedy decoding given a context.
This is of more than theoretical interest from a language modeling perspective. 
Language models trained using the standard cross-entropy loss function are in effect being optimized to perform this very task,
and furthermore, many NLP applications rely in practice on effective and robust word prediction. 

A standard and widely-used metric for evaluating language model performance is with \emph{perplexity} ($PPX$), which is closely related to this prediction task.
When computed for a given token prediction event by a language model, $PPX$ captures how ``predictable'' that event was for the model:

\begin{equation} \label{eq:ppx}
PPX(p,q)=-\sum_{X}p(x)\log q(x)
\end{equation}

Where $X$ corresponds to $V$ (the model's vocabulary of possible tokens it must choose between), $p(x)$ represents the ``true'' or ``target'' distribution and $q(x)$ the model's estimated distribution. 
The closer the predicted distribution matches the target distribution, the lower the perplexity. 
When averaged over many prediction events, and computed on a held-out test dataset, perplexity attempts to capture the degree to which the model has optimally learned to represent its target distribution.
A more accurate (i.e., ``better'') model should result in lower average perplexity (as the model will more often predict a high probability for the correct target).

\subsection{Evaluation Considerations}
\label{subsec:considerations}

Perplexity is a classic example of an \emph{intrinsic} evaluation metric, in that it is measuring the model's ability to carry out its immediate objective.
As mentioned previously, modern language models are often evaluated according to their performance when used as components in a downstream task of some kind.\footnote{\citet{galliers1993} refer to this as the model's ``function'' (in contrast  to its ``objective.'')}
We find this increasing prevalence of \emph{extrinsic} evaluation to be a very positive development, and do not in any way wish to argue \textit{against} use of downstream tasks for evaluation.
However, we see several limitations to an extrinsic-only evaluation paradigm, and argue for more robust intrinsic measures.\footnote{In this, we follow~\citet{ito1999new}, who, writing about language models in the context of their use in ASR systems, warned against relying solely on evaluation metrics that were specific to that task (specifically, word error rate).} 
Extrinsic evaluation is necessarily dependent on the selection of specific benchmark tasks to include, and this process is fraught with difficulty, for several reasons.
First, there are many possible benchmark tasks from which one could choose, each attempting to measure something different.
Different authors will naturally choose different combinations of tasks when evaluating their language models, as they may be focused on different aspects of their models' behavior. 
While scientifically appropriate, this does make for a heterogeneous evaluation landscape, and complicates comparisons between published results.
Second, new tasks are constantly being created, and existing tasks are regularly updated. 
This results in a complex and unstable evaluation landscape in which evaluation tasks change from year to year, and allows for much confusion around versions of datasets and benchmarks. Third, downstream NLP tasks and datasets often have their own issues around validity. For example, the commonly-used SNLI natural language inference corpus~\citep{bowman-etal-2015-large} was later found to have substantial issues resulting from artifacts in how its annotations were collected~\citep{gururangan-etal-2018-annotation}.
How should one now assess a language model evaluated using this downstream task, knowing that the metrics may be of very limited validity?
Finally, we note that widely-used and well-studied downstream evaluation tasks are often not available in ``low-resource'' languages, and so may not be an option in many scenarios.
For these reasons, we believe that intrinsic measures should still play an important role in language model evaluation.

The question then becomes that of \emph{what} to measure.
Perplexity has the advantage of being well-understood and easy to calculate, and is closely linked to the standard cross-entropy training loss frequently used in language modeling.
However, it has long been observed that perplexity itself often fails to correlate with downstream task performance~\citep{iyer1997,ito1999new}, suggesting that it may have limited external validity as a metric.

There is an additional, more subtle limitation to the use of perplexity in cross-model comparison. 
As previously mentioned, many modern language models use sub-word units of prediction.
One of the consequences of this heterogeneity is that evaluation metrics that relate to individual base-level prediction events (as is the case with perplexity) are not comparable across models, even if they are trained and evaluated on the same corpus: different tokenizations and vocabularies will result in different numbers of prediction events, as well as a differently-sized space of possible choices at each event.
From the perspective of the perplexity metric, two models with different approaches to tokenization are performing fundamentally different and numerically incomparable tasks.

Beyond this statistical problem, there is a problem with the underlying semantics of using perplexity as a measure when working with sub-word units.
Any actual application of a language model that involves explicit word prediction\footnote{For whatever definition of ``word'' is appropriate in the language under consideration.} will ultimately demand not \emph{fragments} of words, but rather \emph{entire} words.
In other words, even models whose native unit of prediction is at the sub-word level must make predictions that can \emph{eventually} be able to be decoded into whole words at \emph{some} point. 

Given that, raw perplexity becomes a somewhat confusing evaluation metric, as the underlying phenomenon that it is measuring is quite distinct from the model's actual objective (i.e., predicting a whole word).
Imagine, for instance, a model that predicts at the sub-word level, and now must predict a word given the history ``\emph{The tyrannosaurus was chased by the}.''
The correct continuing word is ``\emph{velociraptor},'' and under the sub-word tokenization used by this model, this will necessitate several separate prediction events (as ``velociraptor'' is both a long and an infrequently-occurring word).
From the perspective of the perplexity metric, however, there will be no difference between the first unit or the third.\footnote{Or, for that matter, from the previous token, ``\emph{the}.''}
Whatever the perplexity metric is telling us about the model's behavior during this process will likely tell us little about the model's ability to actually predict ``velociraptor'' given this particular word history.

\subsection{Recentering on Words}
\label{subsec:refocusing}

We propose that intrinsic evaluation of language models be done in terms of the \emph{whole-word} prediction task, regardless of the specific tokenization practices of any particular model.
This would have the advantage of making cross-model comparison easier, and of the resulting metric bearing a closer resemblance to what we intuitively expect such a metric to capture (i.e., the model's performance at its primary objective).
While computing perplexity at the level of whole words (see section~\ref{sec:methods}) is a step in the right direction, we also propose several additional intrinsic metrics relating to the word prediction task.

\textbf{Word Prediction Accuracy} 
We propose directly measuring and reporting the model's raw \emph{accuracy} at word-level predictions (i.e., the proportion of words that were predicted correctly). 
This has the advantage over perplexity of grounding the number more closely to the concrete performance objective that we are concerned with.
Furthermore, it is easily extended to account for various attributes of model behavior that may be of interest in terms of downstream tasks, while still remaining in the realm of intrinsic evaluation.

In the experiments we describe in section~\ref{sec:methods}, we experiment with variations on this metric that capture different notions of ``accuracy.''
For example, we explore ``top~$n$'' accuracy (i.e., if the target word is in within top $n$ most likely predictions, that prediction counts as a ``hit'').
This could be of use in a text entry scenario, in which the model is responsible for generating candidate words for further selection or refinement by an end-user (as in a mobile phone keyboard application).
Many other possible downstream tasks for language models involve  techniques that would also benefit from having the target word given better placement in the ranked prediction space, and thus would benefit from a metric explicitly measuring this property.

\textbf{``Soft-Match'' Prediction Accuracy}
We propose extending simple prediction accuracy to allow for ``near miss'' predictions, where the predicted word is ``similar'' to the target (for a specified definition of ``similar'').
In many applications of language modeling, there may be multiple possible valid predictions.
This problem has long been understood in the context of machine translation evaluation; in their description of the motivation behind the METEOR metric, \citet{lavie2009meteor} addressed the ``problem of reference translation variability by utilizing flexible word matching, allowing for morphological variants and synonyms to be taken into account as legitimate correspondences.''
In a word prediction task, we could allow an explicit synonym to count as a correct prediction; depending on the application or domain in question, one could use external language resources to model much more complex and task-specific notions of similarity (e.g., in a biomedical NLP context, one might give the model credit at evaluation time for predicting a medication that is from the same functional class as the target).
In the experiments described in section~\ref{subsec:methods:softmatch}, we use a method based on word neighborhoods in an embedding space.
Depending on the nature of the task under consideration, other features could be used.

Or, consider a typing task in a morphologically rich language, in which a user might be willing to accept predictions that involve the correct lexeme but with an incorrect inflection.
Allowing for this sort of flexibility in the evaluation of a word prediction model has the potential to greatly increase the ecological validity of the experiment, in that, that the experimenter is able to easily encode their own task-specific notions of relevance while still staying in a fairly constrained and easy-to-analyze evaluation setting.

\textbf{Lexical Frequency \& Diversity}

One important limitation of raw classification accuracy as a metric is its susceptibility to being biased by imbalanced class distributions.
For example, if some classes occur much more frequently than others, a model may achieve a high accuracy score by learning to focus on these frequent classes to the exclusion of infrequent ones.
In written language, the distribution of classes (i.e., of word \emph{types}) are notoriously skewed~\citep{zipf1935}, and exhibit a ``long tail'' of words that occur relatively infrequently, with a small set of ``head'' words that make up a large proportion of individual \emph{tokens} observed in the training and test data.

We observe that language models often exhibit very different performance characteristics when predicting more common types than less common types; in fact, our experiments in this paper demonstrate that, for some commonly-used language models, the actual number of infrequent types that are \emph{ever} correctly predicted is surprisingly small (see section~\ref{subsec:diversity}).
This over-emphasis on frequent types, when carried forward into downstream generation tasks, may lead to the failure mode described by~\citet{holtzman2019curious} in which generated text is ``dull and repetitive.''
This phenomenon is not limited to words alone; morphologically-rich languages (MRLs) exhibit a similar Zipfian distributional pattern in terms of the occurrence of different morphological phenomena, which in turn affects the performance of systems designed to process such features of language~\citep{czarnowska2019don,tsarfaty2020spmrl}.
We believe that this behavior can be explained through the lens of the bias-variance tradeoff common to all statistical learning problems.
As observed by~\citet{lazaridou2015hubness}, neural models have a tendency towards the ``bias'' end of that tradeoff, which in the context of language modeling results in a strong preference for head words and against tail words.

This is a serious enough problem in machine translation and text generation systems that there is a growing body of literature looking at ways to increase the lexical diversity in model output.
Some authors~\citep{li2016diversity,welleck2019neural} have examined training strategies and loss functions that optimize for diverse output, while others~\citep{vijayakumar2016diverse,ippolito2019comparison} focus on alternatives to greedy decoding and identify several ways to generate more diverse sequences of words.
Questions of evaluation arise, as the construct of ``diversity'' itself is surprisingly difficult to characterize, as pointed out by~\citet{tevet2020evaluating}.

In the context of our word prediction task, we propose two evaluation measures that account for the Zipfian skew in type distributions, and illuminate differences in model performance across the type frequency spectrum.
First, we propose \emph{stratifying} our evaluation of prediction accuracy by frequency, such that we separately measure the model's ability to predict occurrences of high-, mid-, and low-frequency types (stratified \emph{token} coverage). 
Second, we propose measuring the \emph{overall proportion of possible types} that the model was able to predict at least once during evaluation (\emph{type} coverage, also stratified by frequency).

\section{Methods}
\label{sec:methods}

In this section we describe a series of experiments in which we use our proposed evaluation metrics to explore the behavior of several widely-used and large-scale language models (obtained using the  HuggingFace~\citep{Wolf2019HuggingFacesTS} Transformers library).
Specifically, we examine GPT-2~\citep{radford2019language} (\texttt{gpt-2}), GPT~\citep{radford2018improving} (\texttt{openai-gpt}), RoBERTa~\citep{liu2019roberta} (\texttt{roberta-base}), and BERT~\citep{devlin-etal-2019-bert} (\texttt{bert-base-uncased}). 

\subsection{Training \& Datasets}
\label{subsec:training}

Since the pre-trained models were all trained in widely varying ways on different corpora, we ran each model through a single pass of fine-tuning on a common corpus to attempt to bring them more closely into alignment.
For this fine-tuning (and for the ensuing experiments), we used WikiText 103~\citep{merity2016pointer}, which consists of a large ($n=28,475$) training set of English-language Wikipedia articles and a small ($n=60$) test set of 60 articles, with one sentence per line.
The fine-tuning task was on a word prediction task in a unidirectional fashion, in which the context is based only past history (i.e., not on future tokens).\footnote{Bert and Roberta were given a `[MASK]` token at the end on a sequence to ensure unidirectional prediction.}
We note that for BERT and RoBERTa, this usage does differ somewhat from the prediction paradigm under which they were trained, which is implicitly bidirectional.

\subsection{Whole-word decoding}
\label{subsec:whole_word}

As previously described, modern language models typically use sub-word/sub-sentences units as their native unit of prediction.
In order to perform a meaningful evaluation of cross-model word prediction accuracy, it is necessary to obtain word-level predictions, which for the mentioned models 
may involve more than one model-level prediction event.
The models we worked with in this set of experiments used two different tokenization strategies (wordpieces for GPT and BERT, and BPE for GPT-2 and RoBerta), and as such we developed algorithms for decoding whole words by sequentially decoding individual sub-word units.
While the algorithms differ slightly in their implementation between the model families, the overall method is similar.

Our single-word decoding algorithm extracts the first word candidate by the model through concatenating tokens until \textit{end-of-word} is indicated,~\footnote{The code is available for all model types we present in this paper, and for the different tokenization approaches by which they are trained.} and then compared with a target word (see App.~\ref{algs}, Algorithm~\ref{argmax_alg}). 
To extract multiple candidate words, given a target word we run a Depth-First Search to find whether a valid path of tokens exist, having each model prediction spanning its top ten guesses (App.~\ref{algs}, Algorithm~\ref{bfs_alg}). 
This is not a typical beam-search based on likelihoods, but rather is based on the existence of valid units (in the first K options) for a given target word, simulating user choices given a context.~\footnote{Code at \url{https://github.com/shiranD/word_level_evaluation}}

In addition to decoding whole words, we would like to be able to obtain a probability estimate of the resulting prediction, for use in computing a word-level perplexity measure.
We approximate this by taking the product of the prediction-level probabilities (i.e., the model's estimate of the probability of each constituent unit in a given decoded word), which we can then use for a perplexity-like score:

\begin{equation} \label{eq:our_ppx}
ppx={-\sum^{words}p(w)\log (\prod^{units} q(u))}
\end{equation}

\subsection{Experiments}

We performed a word-level prediction experiment on the test dataset described in Section~\ref{subsec:whole_word}, using each of the models in Section~\ref{sec:methods}.
For each test example, we performed incremental unidirectional word prediction using Algorithm~\ref{argmax_alg}
to generate whole-word predictions. 
In other words, for each test example $W$ comprised of $w_1 \ldots w_n$ words, we queried the model $n-1$ times, to predict
$
\hat{w_i} = \underset{w}{\operatorname*{argmax}} P(w_i | w_{1:i-1} )
$
for $i \in \left[ 2, n \right]$.
Additionally, we used Algorithm~\ref{bfs_alg}
to decode the top $k$ ranked word predictions (for $k=10$), $\hat{\mathbf{w}}_t^k$.
In other words, for the test input ``the dinosaur ate the ...'' we would sequentially predict $p(w_2 | \text{``the''})$, $p(w_3 | \text{``the  dinosaur''})$, and so on.
At each prediction event, we compared the predicted $\hat{w_t}$ to the ground-truth $w_t$ according to the various metrics described in the next section. We counted as ``hits'' word-level prediction events where the comparison matched (for the different definitions of ``matched''), and ``misses'' otherwise.

\subsection{Calculation of Metrics}
\label{subsec:calculation}

\subsubsection{Prediction Accuracy}

We measure token-level prediction accuracy\footnote{For tokens--- i.e., words--- in the test set, as opposed to tokens from the perspective of the model being evaluated.} using an exact-match criterion, $top_1$. In other words, if $w_t = \hat{w_t}$, a ``hit'' is counted; otherwise, a miss.
We also computed a higher-recall metric $top_k$, in which a ``hit'' is counted if $w_t \in \hat{\mathbf{w}}_t^k$ --- i.e., if the target word is in the top $k$ predictions, it counts as a ``hit.''
For our experiments, we computed $top_{10}$ (i.e., $k=10$).

\subsubsection{Soft-Match Accuracy}
\label{subsec:methods:softmatch}

As described in section~\ref{subsec:refocusing}, there are a number of criteria by which one might implement a soft-matching algorithm.
From the perspective of evaluation, the key is to design a criterion in such a way as to capture the aspect of user behavior that one may wish to support.

We performed our soft-matching experiments with a text entry scenario in mind, in which a user is able to choose among the language model's top $n$ predictions.
Under this scenario, if the model fails to predict the target word but instead predicts a \emph{related} word (a synonym, perhaps), the user may still be able to convey their message. To simulate this, we may define the soft-match operation as follows:

$$
SoftMatch(a,b,s) = 
\begin{cases}
True & \text{if } a = b \\
s(a,b) & \text{otherwise}
\end{cases}
$$

Where the arguments to $SoftMatch$ are two candidate words $a$ and $b$, as well as a similarity function $s:(a,b) \rightarrow X \in \{True, False\}$.
$Softmatch(a,b,s)$ is true if $a$ and $b$ are a match, or if $s$ indicates similarity.
For our experiments here, we used a method based on similarity in word embedding space, under the theory that words with similar embeddings may be (relatively) appropriate substitutions in a word prediction task.

We used the word2vec algorithm~\citep{mikolov2013distributed} to train 50-dimensional word embeddings on the ``train'' subset of the WikiText-103 corpus.
We then defined our softmatch similarity function $s_{knn}(a,b) = a \in knn(b)$, where $knn(b,k)$ retrieves the $k$ nearest neighbors of target word $b$ in the embedding space.
Using our softmatch function, we then re-scored the prediction accuracy such that a positive softmatch counted as a ``hit.''
We used the Annoy library~\citep{Bernhardsson:2018aa} to perform efficient nearest-neighbor retrieval.
We conducted experiments in which we varied the $k$ parameter; in other words, by allowing a match deeper into the $k$-nearest neighbors of the target.
Our motivation for this was that, ceteris paribus, a model that mis-predicts a target but at least guesses something that lies in the right semantic neighborhood is more useful than one that does not.

\subsubsection{Lexical Diversity}

In order to measure type diversity given all the hits in $top_1$/$top_{10}$, we counted how many unique types were correctly predicted for the first and top-ten guesses and present it in $T_1$ $(T_{10})$ respectively. 
To illustrate the utility of measuring the rate of unique types that were correctly predicted, consider a hypothetical dataset in which $20\%$ of the tokens consist of the word \textit{the}, and that the model at hand predicts only this word for every sample in the test set. 
In this scenario, $top_1$ accuracy will be $20\%$, as \textit{the} is a correct prediction for $20\%$ of the times, yet $T_1$ is based on only one type~\footnote{$T_1$ is the relative percentage of one over the overall number of types} as there was only a single type that was correctly predicted---  suggesting sub-optimal learning of the input distribution, or a lack on the model's ability to reflect that distribution during test.

\section{Results}

\subsection{Diversity Evaluation}
\label{subsec:diversity}


\begin{table}[h]
\begin{center}
\begin{tabular}{lrrr}
model &  $top_{1}$ $(top_{10})$ & $T_{1}$ $(T_{10})$ & $ppx$\\ \hline

{\tt GPT-2} & $35.63$ $(67.76)$ & $26.60$ $(47.27)$ & $34.8$\\ 
{\tt GPT} & $29.37$ $(60.89)$ & $15.96$ $(30.80)$ & $37.9$ \\ 
{\tt RoBerta} & $28.18$ $(59.55)$ & $24.73$ $(42.63)$ & $42.2$ \\ 
{\tt Bert} & $22.11$ $(50.98)$ & $15.59$ $(29.61)$ & $50.7$\\ 
\end{tabular}
\end{center}
\caption{\label{wiki} Experimental results on Wiki-103 corpus}
\end{table}

Table~\ref{wiki} describes the results on the different models. {\tt GPT-2}, and {\tt GPT}, that were pre-trained for word prediction task exhibited the lowest $ppx$. {\tt GPT-2} had the highest hit rate, and type diversity. However, when comparing {\tt GPT}, to {\tt RoBerta}, while accuracy seems to present a similar performance, and the $ppx$ is lower for {\tt GPT}, {\tt RoBerta} is found to be much more diverse than {\tt GPT}, suggesting that the similar hit rates ($28.18$, $29.37$) can be attributed to different reasons as shown by their different performance over $T_x$ metric. 
On the other hand, we can also learn that while {\tt Bert}, and {\tt GPT} share similar diversity rate (prediction diversity), {\tt GPT} exhibits a higher prediction accuracy making for a different accuracy/diversity ratio than {\tt Bert}, which may also suggest a different prediction behavior than of {\tt Bert}'s. To understand type diversity we must explore which types were predicted well and which types were harder. To this end, we stratified both $T_1$, and $top_1$ as a function of frequency; \textit{high}, \textit{mid}, and \textit{low}, for $x\in[10^3,\inf)$, $x\in[10^2,10^3)$, $x\in[10^1,10^2)$ where $x$ is each target type's frequency. 

\begin{figure}[ht]
    \centering
    \includegraphics[width=0.47
    \textwidth]{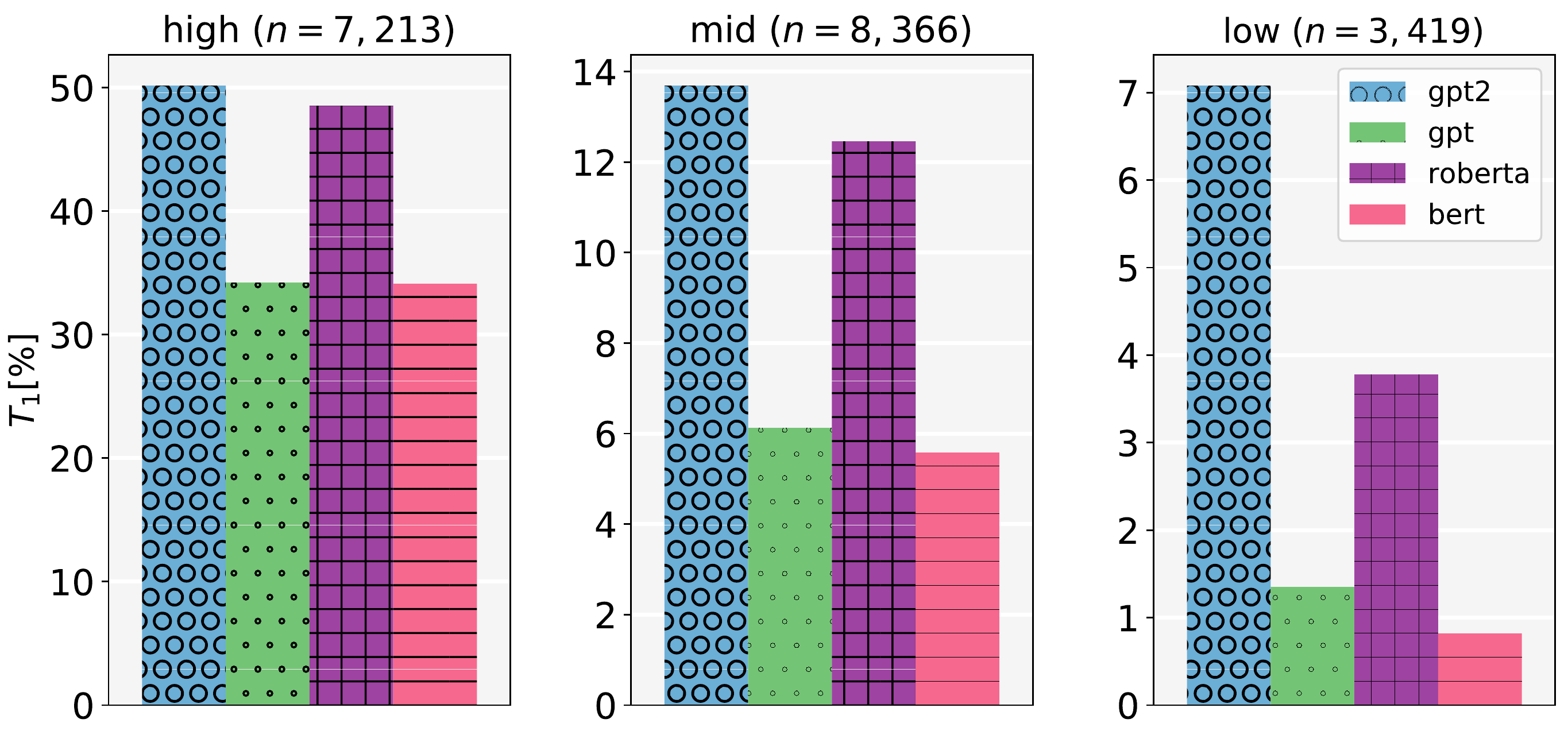}
    \caption{Wiki-103 \textit{type} coverage by training frequency bin. $n$: number of items in each bin; y-axes are percentages over $n$ (note different scales).}
    \label{fig:types_wiki}
\end{figure}

Figure~\ref{fig:types_wiki} describes the type distribution reflecting high diversity for both {\tt GPT-2}, and {\tt RoBerta}, while having {\tt GPT-2} picking on the low-bin twice as many than {\tt RoBerta}. Notice the stark difference between {\tt RoBerta}, and {\tt GPT}, {\tt RoBerta} outperformed {\tt GPT} across every bin, illustrating its diversity strength (given the similar hit rate shown earlier). 
While performing worse, both {\tt GPT}, and {\tt Bert}, seem to share similar rates of diversity, with {\tt GPT}, performing almost twice as many on the lowest bin. Finally, even {\tt GPT-2} that attained the highest diversity, was covering only $50\%$, $14\%$, and $7\%$ of the trained types we evaluated on. This shows there is room for improvement to reflect more optimally the input data's distribution.

\begin{figure}[ht]
    \centering
    \includegraphics[width=0.47
    \textwidth]{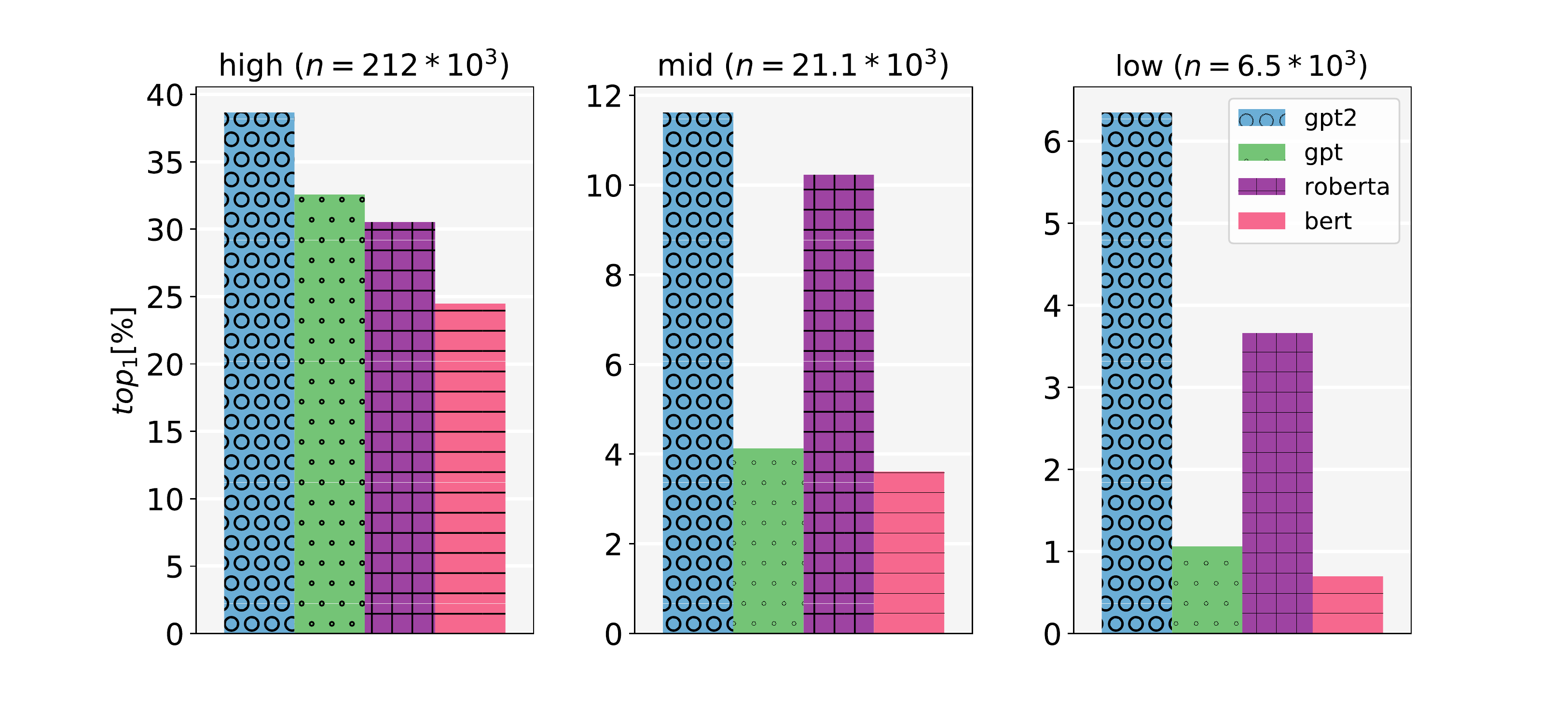}
    \caption{Wiki-103 \textit{token} coverage by training frequency bin. $n$: number of items in each bin; y-axes are percentages over $n$ (note different scales).}
    \label{fig:tok_wiki}
\end{figure}

Figure~\ref{fig:tok_wiki} presents the hit rate distribution. This figure explains the gaps of {\tt GPT-2}, and {\tt RoBerta}, showing that while not so different in diversity, {\tt RoBerta} is missing the hits mostly from the most frequent bin $10\%$ gap, and a sub-optimal prediction in the mid- and low-bins. The similar hit rate of {\tt RoBerta}, and {\tt GPT}, clearly is distributed differently having {\tt RoBerta} reaching parts of the long tail of the distribution more often than {\tt GPT}. {\tt Bert}, and {\tt GPT}, also exhibit the biggest gap in the most frequent bin with $8\%$ difference, while the mid and low bins are similar. Overall evaluating prediction diversity can inform us about the model's priorities. Through measuring type diversity, we learn that models that share similar hit rates, can be vary immensely in diversity, which later on may impact downstream tasks. Evaluating diversity could not only inform us to what degree the learned distribution is reflected, but could directly point at the missing types, and the weaknesses of the model. Since all these models are shown to be weaker in the lower bins, or biased by frequency, our community can benefit if we start addressing this problem, which indirectly would contribute to higher accuracies as well. 
In Section~\ref{para} we illustrate in a case study why learning diverse types, and low-frequency types in particular can be useful. Next, we present a way to further understand our models, even if the target word was not found directly in a prediction.

\subsection{Soft-Match Evaluation}
\label{subsec:softmatch}

\begin{figure}[ht]
    \centering
    \includegraphics[width=0.47
    \textwidth]{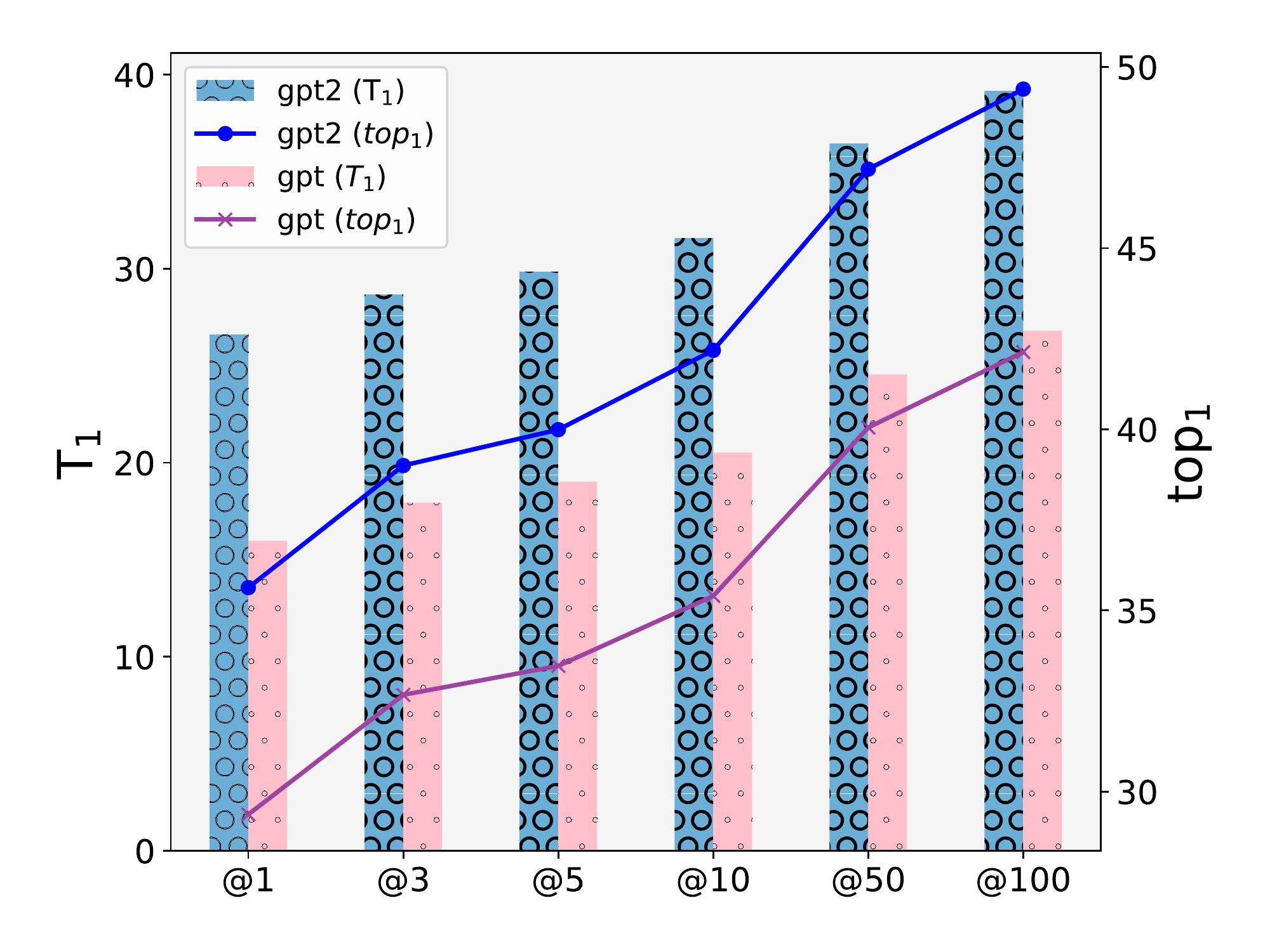}
    \caption{Soft-Match for {\tt GPT-2}, and {\tt GPT} (Wiki-103)}
    \label{fig:soft_match}
\end{figure}

Figure~\ref{fig:soft_match} illustrates {\tt GPT-2}, and {\tt GPT}'s $T_1$, and $top_1$ performance on left (bars) and right (line) axes. Both models gradually ($@3$-$@100$) capture more types as the beam of $k$ in $knn$ was increased (considering more target-neighbors), leading to increased hits. This evaluation shows that {\tt GPT-2} exact match ($@1$) are higher, but that its misses can enrich the pool of unique types with $14\%$ ($@100$) additional unique types (light blue), whereas, {\tt GPT-2} covered only $11\%$ more types (light pink), while both models increase in accuracy is similar. This reinforces that the models' prediction mechanism is slightly different, as similar gains in accuracy are translated to either more of learned high types or more diverse patterns shown for the models in Figures~\ref{fig:types_wiki},~\ref{fig:tok_wiki}. 
This analysis teaches us that even if there were mis-matches some of them were actual near misses, and are related to what it was expected to predict, which as mentioned can be of practical use for different users, or for analyzing how wrong were the mis-matches as part of an error analysis process.

\section{A case study of Paraphrasing}\label{para}
In this section we will look at the impact of model inference performance on the particular downstream task of paraphrasing. To this end we employed a SotA algorithm, Bertscore~\citep{zhang2019bertscore}, to compute similarity scores of sentence pairs in part by comparing embeddings derived from a language model. Under Bertscore, higher similarity scores indicate greater semantic similarity of a pair of sentences, such that one is a closer \textit{paraphrase} of the other. We would like to stress that the critique that may be risen at the end of this section is not about Bertscore tool as such, but are rather about a certain type of pattern that the models that are employed by this tool may have insufficiently learned. 

\textbf{Why Paraphrasing?} We choose the downstream task of paraphrasing to measure semantic similarity of sentence pairs as it can be easily manipulated to consider a single word modification. Consider the following example sentence involving the word triple: (poodle, dog, cat)
\footnote{Note that the word {\tt poodle} occurs much less frequently in English than either {\tt dog} or {\tt cat}.}

\begin{enumerate}[label=(\alph*)]
\item which \texttt{dog} has longer hair ?
\item which \texttt{cat} has longer hair ?
\item which \texttt{poodle} has longer hair ?
\end{enumerate}

The pair $(a,b)$ ought to score lower (i.e., be considered by Bertscore to be more dissimilar) than the pair $(a,c)$, as $a$ is a valid paraphrase of $c$ while $b$ is not. 
This, of course, assumes that the language model being used as the underlying source of embeddings for the Bertscore algorithm has accurately captured the semantic meaning of the three words under consideration. 
If not, we may see an inversion of results such that $(a,b)$ appears (incorrectly) more similar than $(a,c)$, suggesting that the model in question should perhaps not be used for paraphrase-related tasks.


To explore the impact of word frequency on model representations with our fine-tuned models, we have generated $50$ rare, and $50$ common triples elicited from wiki-103 trainset. 
Each of the triples contains a rare/common word, its hypernym, and a sibling hypernym extracted from WordNet~\citep{miller1995wordnet} (using {\texttt {nltk}}~\citep{loper2002nltk}) $(x_r, x_h, x_a)$, $(x_c, x_h, x_a)$ respectively. 
For each word in a triple, we identified a sentence in which the rare word naturally occurs, and generated probe sentences in which  we replace the rare word with $x_h$ and $x_a$
~\footnote{see Appendix~\ref{protocol} for details on sentence selection and generation.}. 
We then used $bertscore$ to compare our sentences in terms of their similarity.
In principle, we expect that $bertscore(s(x_h), s(x_r)) > bertscore(s(x_h), s(x_a))$. 
In other words, a similarity score for the pair made of a sentence with \texttt{dog} and the sentence with \texttt{poodle} is expected to be higher than than the pair made of a sentence with \texttt{dog} and a sentence with \texttt{cat}, as \texttt{dog} and a \texttt{poodle} are closer semantically, than a \texttt{dog} and a \texttt{cat}, and therefore would be a closer paraphrase of each other.
Alternatively, if the model's word representations are being confounded by lexical frequency, we may instead observe the opposite pattern (i.e., the sentences with the more common words mistakenly appearing to be more similar to one another, despite their semantic difference).
We consider cases in which the model correctly identifies the paraphrase (e.g., if $bertscore(s({\texttt{dog}}), s({\texttt{poodle}})) > bertscore(s({\texttt{dog}}), s({\texttt{cat}}))$) as hits, and mis-identifications as misses.
Our ``null hypothesis'' is that there should not be any difference in hit rate between high- and low-frequency words (i.e., word frequency should not affect the model's ability to identify paraphrases).
Furthermore, we compare the performance of two fine-tuned models trained on wiki-103, {\tt Bert} and {\tt RoBerta}, and (given the results of our earlier experiments) we hypothesize that if there is a difference in hit rate, {\tt RoBerta} will prove more robust to the rare words condition, given its superior performance at predicting (and thus representing) rare words.


\begin{table}[h]
\begin{center}
\begin{tabular}{lrrr}
model &  $hits$ & $misses$ & $total$\\ \hline
{\tt Bert}$_{rare}$ & $14$  & $36$  & $50$\\ 
{\tt RoBerta}$_{rare}$ & $11$  & $39$ & $50$  \\ 
{\tt Bert}$_{common}$ & $40$  & $10$   & $50$\\ 
{\tt RoBerta}$_{common}$ & $39$  & $11$ & $50$ \\ 
\end{tabular}
\end{center}
\vspace{-0.15in}
\caption{\label{wiki_para} Paraphrasing sentences with wiki-103 words}
\end{table}

We note that this is something of a ``toy'' experiment, given its small size, which limits the conclusions that we can draw.
However, in Table~\ref{wiki_para}, we do see a greater difference in performance between the rare set of words to the common, such that the models do appear to be failing to capture the semantics of rare words, as reflected in the greater number of $misses$ ($\chi^2$; $p<0.001$ for both models). 

We found that {\tt RoBerta} and {\tt Bert} did not differ greatly in their performance, suggesting that with this method the strong effect of word frequency outweighed the between-model difference observed in our earlier experiments. 
This null result could easily be an artifact of our very small sample size of 100 probe sentences, though, and we also did notice a substantial number of $misses$ with the set of common words.
Overall, despite this being based only on a small sample, it does seem that the lower performance of both models on the rare words is unlikely to be a coincidence.
We hope to be able to experiment with a greater sample size to begin learning more about the degree to which rare-word inferences are reliable to produce outcomes aligned with human semantics on various downstream tasks.

\section{Future Work}
The paraphrasing task should be conducted at a larger scale.
Furthermore, we hope to continue evaluating language models' prediction diversity and its effects on additional downstream tasks (for example, tasks where human speech is anticipated), since prediction diversity evaluation may vary between one task to another. 
The unit of evaluation can go beyond words, and may be defined at various textual granularities, such as phrases, for instance, depending on prediction diversity desired. 
We also leave for future work questions of to what degree different tokenization approaches, or model size effect prediction diversity. 

In our experiments, we did observe differences in performance between models with different tokenization strategies (e.g. {\tt GPT-2} and {\tt RoBerta} as compared with their architectural counterparts); however, these models also varied substantially from one another in other respects (size, etc.) and as such it is difficult to attribute this performance gap to tokenization alone.
It may also be the case that the bigger the model (in terms of number of parameters), the more diverse it is likely to be; under this hypothesis, we would expect today's ever-larger models (e.g. {\tt GPT-3}) to outperform their predecesors in terms of diversity. 
However, we do not believe that it is sustainable~\citep{strubell2019energy,schwartz2019green} to rely on increasing model complexity as an approach to addressing the frequency-related challenges that we observed in our experiments, and believe that fundamentally different approaches to language model training are needed. 


\section{Conclusion}
We presented two types of evaluation techniques to learn about the performance of the model across its input distribution, revealing the easier and the challenging areas to learn. 
Through this analysis we showed that current models are susceptible to frequency bias during training, and as a result under-performing when less frequent examples are encountered at test time, hurting the overall performance. 
In addition, we proposed a way 
to learn about the degree to which a model's prediction is semantically close to a target in cases where an exact match was not predicted, which may more accurately reflect a model's usefulness. 
Thirdly, we showed how a downstream task of paraphrasing may be rendered less reliable, as the models employed struggle to produce semantically-useful representations when rare words are involved.

We believe that language models should reflect the trained distribution more optimally than what we observed in our evaluation, and we should recognize their bias to frequency - making them unfair towards some words, and potentially harmful for our downstream tasks. 
We also believe it is important to take part in setting benchmarks for models' diversity. Finally, distributional representation goes beyond words, and we hope to address more complicated representational tasks as well.

\section*{Acknowledgments}

We thank the anonymous reviewers for their insightful comments and suggestions. This work was supported by the National Institute on Deafness and Other Communication Disorders of the National Institutes of Health under award number R01DC015999.

\newpage

\bibliography{emnlp2020}
\bibliographystyle{acl_natbib}

\appendix
\section{Stratified Bins}
The bin's assignment is based on the words' frequency of the trainset, but the bins can only be based on the intersection of the high-freq words in train, and all the words in test set. Any high/mid/low-freq train word that occurs in the test will be assigned to its appropriate bin. Code is provided. 

\section{Algorithms for computing first and first ten guesses}~\label{algs}

\begin{algorithm}
  \caption{Top1 Target Word Search}
  \begin{algorithmic}[1]
  \Procedure {TargetFind1} {$w_{<t}$, $w_t$, $model$}
    \State $cxt \leftarrow w_{<t}$
    \State $word \leftarrow empty$
            \While{\varParamE{($word$)} in \varParamE{($w_t$)}} 
                \State $cxt \leftarrow \varParamF(cxt,word)$
                \State $P_{wpc_t} \leftarrow \varParamA(cxt, model)$
                \State $top_{1} \leftarrow \varParamG(P_{wpc_t})$
                \State $word \leftarrow \varParamF(word,top_{1})$
                \If{\varParamE{($word$)} = \varParamE{($w_t$)}} 
                    \State return True
                \EndIf
                \If{\varParamE{($word$)} not in \varParamE{($w_t$)}} 
                    \State return False
                \EndIf
            \EndWhile
    \EndProcedure
  \end{algorithmic}\label{argmax_alg}
\end{algorithm}

\begin{algorithm}
  \caption{TopK Depth-First Search}
  \begin{algorithmic}[1]
  \Procedure {TargetFind2} {$w_{<t}$, $w_t$, $model$}
    \State $cxt \leftarrow w_{<t}$
    \State $P_{wpc_t} \leftarrow \varParamA(cxt, model)$ 
    \State $top_{10} \leftarrow \varParamB(P_{wpc_t})$ 
    \For{$root$ in $top_{10}$}  
        \State $roots.append(root)$
    \EndFor
    \For{$root$ in $roots$}
        \State $paths \leftarrow \varParamC(root)$
        \While{$paths$} 
            \State  $path \leftarrow \varParamD(paths)$
            \State $basic_{cxt} \leftarrow w_{<t}$
            \State $cxt \leftarrow \varParamF(basic_{cxt},path)$ 
            \State $P_{wpc_t} \leftarrow \varParamA(cxt, model)$
            \State $top_{10} \leftarrow \varParamB(P_{wpc_t})$
            \For{$wpc_t$ in $top_{10}$} 
                \State $word \leftarrow \varParamE(wpc_{pre},wpc_t)$
                    \If{$word$ = \varParamE{($w_t$)}} 
                        \State return True 
                    \ElsIf{$word$ in \varParamE{($w_t$)}} 
                        \State $new \leftarrow \varParamF(path,wpc_t)$
                        \State $paths.append(new)$
                    \EndIf
            \EndFor

            \EndWhile

    \EndFor
    \State return False
    \EndProcedure
  \end{algorithmic}\label{bfs_alg}
\end{algorithm}

\section{Protocol to elicit paraphrase pairs}\label{protocol}
We provide here the guidelines to elicit rare/common paraphrase sentences. 
First, a simple routine extracts possible triples that constructs the linguistic relationship desired of hypo/hypernym and a hypernym-sibling ($(x_r, x_h, x_a)$).
We begin by taking the intersection of the vocabulary seen in the training partition of the Wiki-103 corpus with that found in WordNet (using the NLTK package~\citep{loper2002nltk}), and then further filtered for vocabulary items with WordNet entries exhibiting the desired linguistic relationship.
(A synonym/antonym construction could also have been chosen alternatively). 

The frequency dynamic for the rare/common triples $(x_r/x_c, x_h, x_a)$ was (low, mid/high,mid/high) and for the common (mid/high, mid/high, mid/high) respectively.
Words occurring fewer than 50 times in the Wiki-103 training partition were categorized as ``low,'' and were categorized as ``mid/high'' otherwise.
Finally, a human annotator manually identified an appropriate context sentence for each target word via online search across the following web-based dictionaries: ~\url{merriam-webster.com}, ~\url{thesaurus.com},
~\url{sentencedict.com}, and ~\url{dictionary.cambridge.org}.
Here is a rare-word triple example following $(x_r, x_h, x_a)$ order. 
\begin{enumerate}[label=(\alph*)]
\item The \texttt{afterdamp} occurring in such situations is a mixture of carbon dioxide and carbon monoxide.
\item The \texttt{gas} occurring in such situations is a mixture of carbon dioxide and carbon monoxide.
\item The \texttt{liquid} occurring in such situations is a mixture of carbon dioxide and carbon monoxide.
\end{enumerate}
Here is a common-word triple example following $(x_c, x_h, x_a)$ order \begin{enumerate}[label=(\alph*)]
\item Because of the poor economy, the factory will immediately \texttt{discontinue} operations.
\item Because of the poor economy, the factory will immediately \texttt{cease} operations.
\item Because of the poor economy, the factory will immediately \texttt{continue} operations.
\end{enumerate}
The complete sentence list can be found at \url{https://github.com/shiranD/word_level_evaluation}.




\end{document}